\documentclass[sigconf]{acmart}
\AtBeginDocument{%
  }

\copyrightyear{2026}
\acmYear{2026}
\setcopyright{cc}
\setcctype{by}
\acmConference[MM '26]{Proceedings of the 34th ACM International Conference on Multimedia}{November 10--14, 2026}{Rio de Janeiro, Brazil}
\acmBooktitle{Proceedings of the 34th ACM International Conference on Multimedia (MM '26), November 10--14, 2026, Rio de Janeiro, Brazil}
\acmDOI{10.1145/3767308.3836148}
\acmISBN{979-8-4007-2213-4/2026/11}

\begin{document}

\title{GarmentWeaver: Schema-Aware Structured Synthesis for Multimodal Sewing Patterns}

\author{Yinwen Lu}
\authornote{Both authors contributed equally to this research.}
\affiliation{%
  \institution{Donghua University}
  \department{College of Textiles}
  \city{Shanghai}
  \country{China}}
\email{1249003@mail.dhu.edu.cn}

\author{Weihao Luo}
\authornotemark[1]
\affiliation{%
  \institution{Ningbo University}
  \department{College of Science and Technology}
  \city{Ningbo}
  \state{Zhejiang}
  \country{China}}
\email{luoweihao@nbu.edu.cn}

\author{Yueqi Zhong}
\correspondingauthor
\affiliation{%
  \institution{Donghua University}
  \department{College of Textiles}
  \city{Shanghai}
  \country{China}}
\email{zhyq@dhu.edu.cn}
\renewcommand{\shortauthors}{Lu, Luo and Zhong}

\begin{abstract}
Multimodal Sewing pattern generation aims to infer executable sewing patterns from design cues such as sketches and textual descriptions. As an interpretable and simulation-compatible representation, sewing patterns are particularly valuable for digital garment creation. However, existing methods often model garment specifications as flat long sequences, which entangles garment structure with detailed parameters and leads to redundant components, inaccurate local details, and poor simulation compatibility. In this paper, we present GarmentWeaver, a schema-aware framework for multimodal Sewing pattern generation. GarmentWeaver constructs compact hierarchical targets by activating garment-relevant structural branches and predicts executable Sewing patterns in a structured manner. Specifically, we introduce a schema-aware target construction strategy, build the generator on top of a pretrained vision-language model for multimodal garment understanding, and impose feasibility-aware regularization to encourage structurally valid and simulation-compatible outputs. Extensive experiments show that GarmentWeaver produces more accurate and more executable sewing patterns than strong baselines, while also yielding better simulation results. These findings demonstrate the effectiveness of schema-aware structured generation for reliable multimodal Sewing pattern prediction.
\end{abstract}



\begin{teaserfigure}
  \includegraphics[width=\textwidth]{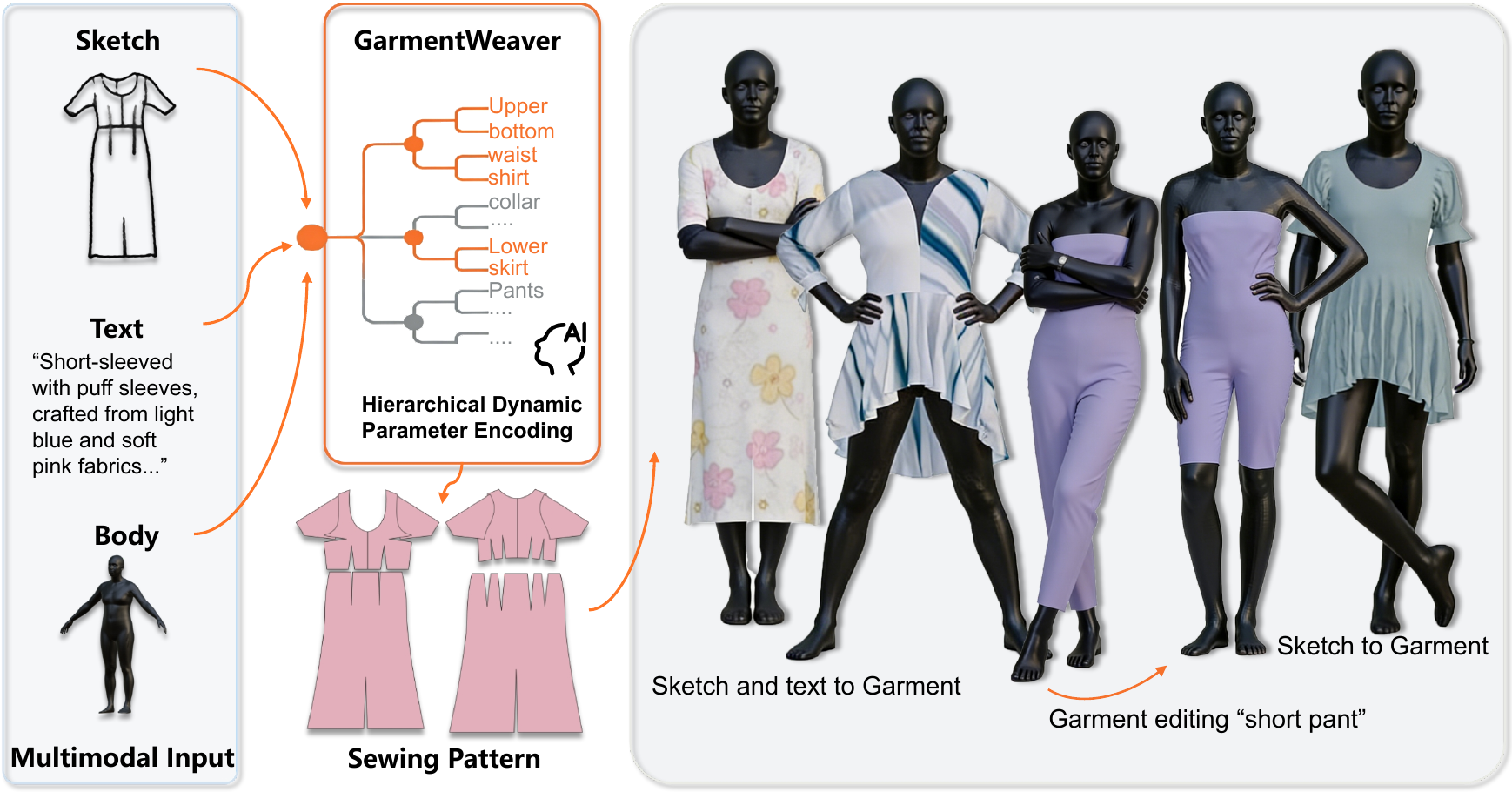}
  \caption{GarmentWeaver generates structured sewing patterns under multimodal conditions, including garment sketches, textual descriptions, and body-related inputs. Through schema-aware hierarchical modeling, the generated patterns preserve structural validity and support downstream garment simulation and editing, enabling controllable digital garment creation from design concepts to simulation-ready outputs.}
  \Description{Enjoying the baseball game from the third-base
  seats. Ichiro Suzuki preparing to bat.}
  \label{fig:1}
\end{teaserfigure}

\maketitle

\section{Introduction}
Garment digitalization has become increasingly important in multimedia applications such as virtual try-on, digital fashion design, gaming, and embodied content creation. Among these tasks, generating structured Sewing patterns from multimodal design inputs is particularly valuable, since sewing patterns provide an interpretable and executable representation that directly supports downstream garment simulation and editing. Compared with image-level or mesh-level generation, structured sewing patterns offer a more practical route from design concepts to simulation-ready digital garments. Recent works have started to move from garment reconstruction toward structured sewing-pattern prediction, while newer methods further incorporate text, sketches, and large multimodal models for more controllable garment generation.\cite{1,2,3,4,5,6,7,8,9,10,11,12,13,14,15,16,17,23,24,25,26,27,28,29,30,31}

Despite this progress, generating valid sewing patterns for complex garments remains challenging. A major reason is that garment specifications are inherently hierarchical: high-level structural choices determine which garment components are present, while lower-level parameters specify the detailed shape of those components. However, many existing methods still model garment outputs in a relatively flat prediction space \cite{1,2,3,4,5,28,30,31}, which forces the model to predict garment structure and detailed parameters in a single long sequence. This often leads to redundant components, inaccurate local details, and poor simulation compatibility. Recent methods such as Design2Garment \cite{3} and ChatGarment \cite{2} have shown the value of program-like or language-friendly garment representations, while AIpparel \cite{1}, SewingLDM \cite{4}, GarmageNet \cite{30}, and GarmentDiffusion \cite{31} demonstrate the promise of multimodal large-model and latent generative formulations. Still, more explicit structure-aware modeling for Sewing pattern generation remains underexplored.

This issue becomes even more evident in multimodal Sewing pattern generation. Sketches provide strong cues about silhouette and contour, while textual descriptions offer complementary semantic information about garment type and design intent. A desirable model should jointly exploit these heterogeneous signals to recover not only the correct garment category, but also fine-grained structural details and executable sewing patterns. In practice, however, direct full-sequence generation often struggles to satisfy these requirements simultaneously, because structural decisions and detailed parameter prediction are tightly coupled. As a result, the model may generate outputs that look locally plausible but still contain redundant branches, mismatched parts, or unstable details that degrade downstream simulation.

To address this issue, we propose \textbf{GarmentWeaver}, a schema-aware framework for multimodal Sewing pattern generation. Instead of directly predicting a single long garment sequence, GarmentWeaver adopts a two-stage structured generation strategy: it first predicts a garment structure template and then fills in the corresponding design parameters. To support this process, we introduce a schema-aware target construction strategy that activates only garment-relevant branches and yields compact hierarchical supervision. We further introduce feasibility-aware regularization to suppress invalid branches and out-of-range parameters, improving the structural validity and simulation compatibility of the generated outputs. Built on top of a pretrained vision-language model, GarmentWeaver effectively integrates sketch, text, and body-related inputs and produces more accurate and more executable sewing patterns than strong baselines.

Extensive experiments demonstrate the effectiveness of GarmentWeaver. Qualitative results show that our method generates more accurate garment categories and finer structural details than representative baselines, while maintaining valid patterns for simulation. Quantitative comparisons further verify improvements in panel reconstruction accuracy, structural correctness, and simulation success rate. In addition, ablation studies confirm that sketch guidance, structured target construction, and feasibility-aware regularization are all important for reliable Sewing pattern generation.

Our contributions are summarized as follows:
\begin{itemize}
\item We reformulate multimodal Sewing pattern generation as a structure-aware prediction problem, where hierarchical garment structure is modeled explicitly instead of being implicitly entangled in a flat long sequence.
\item We propose a schema-aware two-stage generation paradigm that first predicts a structure template and then instantiates its corresponding parameters, yielding compact hierarchical supervision and reducing the ambiguity of direct full-sequence generation.
\item We introduce feasibility-aware regularization to constrain the predicted parameters within structure-valid branches and admissible ranges, improving the structural validity and simulation compatibility of the generated sewing patterns.
\end{itemize}

\section{Related Work}

\subsection{Garment Reconstruction and Sewing Pattern Estimation}

Garment modeling has long been studied in computer vision and graphics. Early works mainly focus on recovering garment geometry or clothed human shape from images, videos, or body-aware priors, such as DeepGarment \cite{9}, Multi-Garment Net \cite{10}, BCNet \cite{11}, TailorNet \cite{12}, and Physics-Inspired Garment Recovery \cite{13}. Later methods further improve geometric reconstruction and draping quality through stronger deformation or implicit priors, including DrapeNet \cite{14}, DIG \cite{15}, ISP \cite{16}, Garment Recovery with Shape and Deformation Priors \cite{17}, and clothed-human reconstruction methods such as PIFu \cite{18}, PIFuHD \cite{19}, ICON \cite{20}, ECON \cite{21}, and Registering Explicit to Implicit \cite{22}. More recent works continue to improve real-world garment reconstruction and simulation readiness through stronger geometry-aware or generative priors, such as CloSe \cite{37}, 4D-DRESS \cite{38}, SIFU \cite{43}, Diffusion-FOF \cite{44}, Gaussian Garments \cite{59}, and reconstruction methods with guided shape and deformation priors \cite{60}. These methods substantially advance garment geometry recovery, but their primary goal is reconstruction rather than executable sewing-pattern generation.

A related line of work moves toward sewing pattern estimation. NeuralTailor \cite{23} reconstructs sewing pattern structures from 3D garment point clouds, while data-driven pattern estimation from 3D geometries \cite{24}, Computational Pattern Making from 3D Garment Models \cite{25}, and single-image sewing pattern reconstruction \cite{26} further strengthen the connection between garment surfaces and their underlying patterns. Other methods such as MulayCap \cite{41}, SewFormer \cite{26}, and more recent in-the-wild sewing-pattern generation approaches \cite{57} also highlight the importance of pattern-aware representations for garment understanding. However, these approaches still mainly treat pattern recovery as estimation from geometry or images, rather than structured multimodal generation from sketch and text. Moreover, they do not explicitly model the hierarchical dependency between garment structure and detailed parameters. 

\subsection{Structured Garment Representations and Program-Based Modeling}

Another important direction is to represent garments in an executable structured form. GarmentCode \cite{6} introduces a programmatic representation for parametric sewing patterns, and GarmentCodeData \cite{7} provides a large-scale dataset of garments paired with sewing patterns, which has become a key benchmark for recent structured garment generation. Other works such as DressCode \cite{5}, Learning a Shared Shape Space for Multimodal Garment Design \cite{27}, GarmentImage \cite{28}, and AutoSew \cite{29} further explore structured, topology-aware, or program-related garment modeling. Design2Garment \cite{3} pushes this direction further by formulating garment creation as a program synthesis problem from multimodal design concepts, while latent flow matching based methods \cite{8} further investigate structured generation in sewing-pattern space.

These works clearly show the value of executable representations for garment generation. However, most existing approaches still model the target as a generic structured sequence, without explicitly accounting for the fact that many garment parameters are valid only under specific structural choices. In contrast, our method adopts a schema-aware formulation in which garment structure is predicted first and detailed parameters are generated only within the corresponding valid branches. This design better matches the hierarchical nature of garment specifications and leads to more compact and more executable outputs.

\subsection{Multimodal Garment Generation with Structured Priors}

Recent progress in multimodal generative modeling has pushed garment generation toward more controllable and practical settings. AIpparel \cite{1} develops a multimodal foundation model for digital garments with a dedicated tokenization scheme for sewing patterns. ChatGarment \cite{2} leverages large vision-language models for garment estimation, generation, and editing, and directly produces language-friendly garment code. Design2Garment \cite{3} also benefits from multimodal design inputs for structured garment creation. In parallel, SewingLDM \cite{4} explores multimodal latent diffusion for complex sewing-pattern generation under text and sketch conditions, while GarmageNet \cite{30} and GarmentDiffusion \cite{31} continue to improve controllability and quality in multimodal garment generation. Recent works such as flow matching based sewing-pattern generation \cite{8}, program-oriented structured synthesis \cite{3}, and topology-aware pattern modeling \cite{28,29} further suggest that structured representations can benefit from stronger multimodal priors.

Beyond garment-specific methods, pretrained multimodal models such as LLaVA \cite{32}, Llama2 \cite{58}, CLIP \cite{46}, and related large-model paradigms for 3D perception and generation \cite{48,49,50,51,52,53} suggest that strong multimodal priors are useful for aligning visual and textual garment semantics. ChatGarment explicitly shows that a VLM can map multimodal inputs to garment programs by leveraging a language-friendly garment representation and simplified GarmentCode-style JSON configurations. However, current garment-generation methods still often rely on flat sequence prediction or latent generation pipelines, which couple structural decisions with detailed parameter prediction in a single output space. This makes them more vulnerable to structural hallucination, redundant components, and non-simulatable outputs. Different from these approaches, our method uses a pretrained vision-language model together with schema-aware structured generation, enabling more reliable prediction of executable Sewing patterns.

\section{Method}

\begin{figure}
  \includegraphics[width=0.5\textwidth]{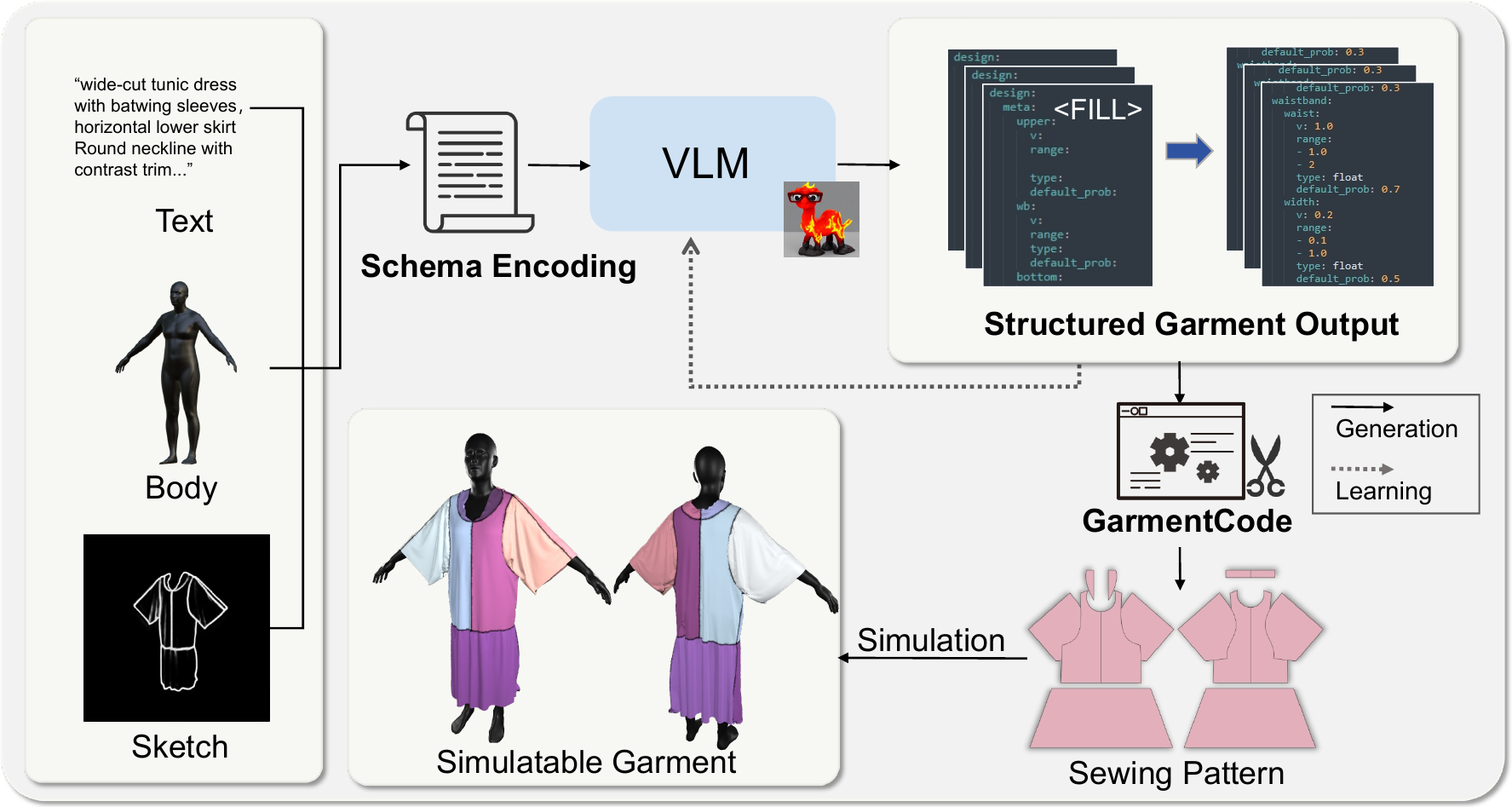}
  \caption{Pipeline of GarmentWeaver. Given text, body, and sketch inputs, the proposed method first constructs a compact hierarchical target through schema-aware dynamic encoding. The VLM then predicts the structured garment output in two stages: it first generates a structure template and then fills in the corresponding design parameters. The predicted output is converted into GarmentCode, from which sewing patterns are constructed and further simulated to obtain a simulation-compatible garment.}
  \label{fig:2}
\end{figure}

We propose GarmentWeaver, a schema-aware framework for multimodal Sewing pattern generation. Given a garment sketch $I$, a textual description $T$, and optional body-related information $B$, the goal is to predict an executable garment representation that can be converted into sewing patterns and further simulated. Unlike conventional flat-sequence generation, GarmentWeaver models garment structure and detailed parameters in a more structured manner, so that the prediction process better matches the hierarchical nature of garment specifications.

\subsection{Problem Formulation}
Let $X=(I, T, B)$ denote the multimodal input and $D$ denote the target garment representation. Existing flat-sequence methods directly model
\begin{equation}
  p(D \mid X)
\end{equation}
which entangles garment structure and detailed parameters in a single long sequence. However, garment representations naturally contain two different types of variables: a structural component $C$, which determines the active garment branches and component configurations, and a parameter component $P$, which specifies the detailed shape under the inferred structure. We therefore decompose the target as
\begin{equation}
  D=(C, P)
\end{equation}
and factorize the conditional distribution as
\begin{equation}
  p(D \mid X)=p(C, P \mid X)=p(C \mid X)\, p(P \mid C, X).
\end{equation}
This factorization separates structure prediction from parameter prediction, reducing long-sequence ambiguity and preventing the model from assigning probability mass to structurally invalid token combinations.

\subsection{Schema-Aware Dynamic Encoding}

A key property of garment representations is that the validity of many parameters depends on garment structure. To encode this dependency, we introduce a schema-aware dynamic encoding. Let $\tilde{P} \in \mathbb{R}^m$ denote the full parameter vector over the complete garment schema. Given a structural component $C$, we define a structure-dependent activation mask
\begin{equation}
  M(C) \in \{0,1\}^m
\end{equation}
and obtain the effective parameter component by
\begin{equation}
  P = M(C) \odot \tilde{P},
\end{equation}
where $\odot$ denotes element-wise masking. Here, $m$ denotes the number of parameter slots in the full garment schema. In this way, only structure-valid fields are preserved, while inactive branches are removed from the target space. Based on the active-branch representation, we further construct a coarse-to-fine target pair:
\begin{equation}
  D^{(1)} = \mathcal{T}(C, P), \quad D^{(2)} = \operatorname{Fill}\left(D^{(1)}, P\right).
\end{equation}

Here, $D^{(1)}$ is a structure template, where valid numerical slots are replaced by placeholder tokens such as \texttt{<FILL>}, and $D^{(2)}$ is the filled garment representation, where the placeholders are instantiated with structure-compatible parameter values. This encoding reduces output redundancy and aligns supervision with the hierarchical structure of garment specifications. In practice, schema-aware pruning also produces substantially more compact targets by removing inactive branches and retaining only structure-valid slots. We provide a quantitative analysis of this compactness in Sec.~4.3.

\subsection{Two-Stage Structured Generation}
We build the generator on top of a pretrained LLaVA-style vision-language model. The pretrained backbone provides strong multimodal features for garment understanding, which help align sketch, text, and body cues during generation.

Given multimodal input $X$, GarmentWeaver performs autoregressive generation in two stages. In Stage I, the model predicts the structure template $D^{(1)}$. This stage focuses on recovering the coarse garment structure, including active branches and placeholder positions. In Stage II, the model conditions on both $X$ and the generated template to complete the parameter component:
\begin{equation}
  p_\theta\left(D^{(2)} \mid D^{(1)}, X\right)=\prod_{t=1}^{L_2} p_\theta\left(d_t^{(2)} \mid d_{<t}^{(2)}, D^{(1)}, X\right).
\end{equation}
Compared with direct one-step generation, this two-stage design better reflects the dependency between garment structure and detailed parameters, resulting in more regular and executable outputs.

The weights $\lambda_{\mathrm{inv}}$ and $\lambda_{\mathrm{range}}$ are empirically chosen to balance the autoregressive objective and the feasibility constraints.

\subsection{Feasibility-Aware Regularization}
The two generation stages are optimized with standard autoregressive objectives:

\begin{equation}
  \mathcal{L}_{\mathrm{temp}}=-\sum_{t=1}^{L_1} \log p_\theta\left(d_t^{(1)} \mid d_{<t}^{(1)}, X\right)
\end{equation}

\begin{equation}
  \mathcal{L}_{\mathrm{fill}}=-\sum_{t=1}^{L_2} \log p_\theta\left(d_t^{(2)} \mid d_{<t}^{(2)}, D^{(1)}, X\right)
\end{equation}
 $L_1$ and $L_2$ denote the sequence lengths of the structure template and the filled garment representation, respectively. $d_t^{(1)}$ and $d_t^{(2)}$ denote the token at step $t$ in the two generation stages. To further improve executability, we introduce a feasibility-aware regularization that constrains the completed representation to remain within the structure-valid space. Specifically, we penalize predictions on inactive branches and out-of-range values on active branches:

\begin{equation}
  \mathcal{L}_{\text {inv }}=\sum_j\left(1-M_j(C)\right)\left\|\hat{P}_j\right\|_2^2
\end{equation}

\begin{equation}
  \mathcal{L}_{\text {range }}=\sum_j M_j(C)\left[\max \left(0, l_j-\hat{P}_j\right)^2+\max \left(0, \hat{P}_j-u_j\right)^2\right]
\end{equation}

where $\hat{P}_j$ is the predicted value of the $j$-th parameter slot and $\left[l_j, u_j\right]$ is its valid schema range. The final loss is

\begin{equation}
  \mathcal{L}=\mathcal{L}_{\text {temp }}+\mathcal{L}_{\text {fill }}+\lambda_{\text {inv }} \mathcal{L}_{\text {inv }}+\lambda_{\text {range }} \mathcal{L}_{\text {range }}
\end{equation}

We set $\lambda_{\mathrm{inv}}=0.1$ and $\lambda_{\mathrm{range}}=0.05$ based on validation performance.By combining schema-aware target construction, two-stage structured generation, and feasibility-aware regularization, GarmentWeaver adapts a pretrained multimodal generator into a structured garment synthesizer that is more controllable, more schema-consistent, and more suitable for downstream simulation.

\section{Experiment}

\begin{figure*}
  \includegraphics[width=1\textwidth]{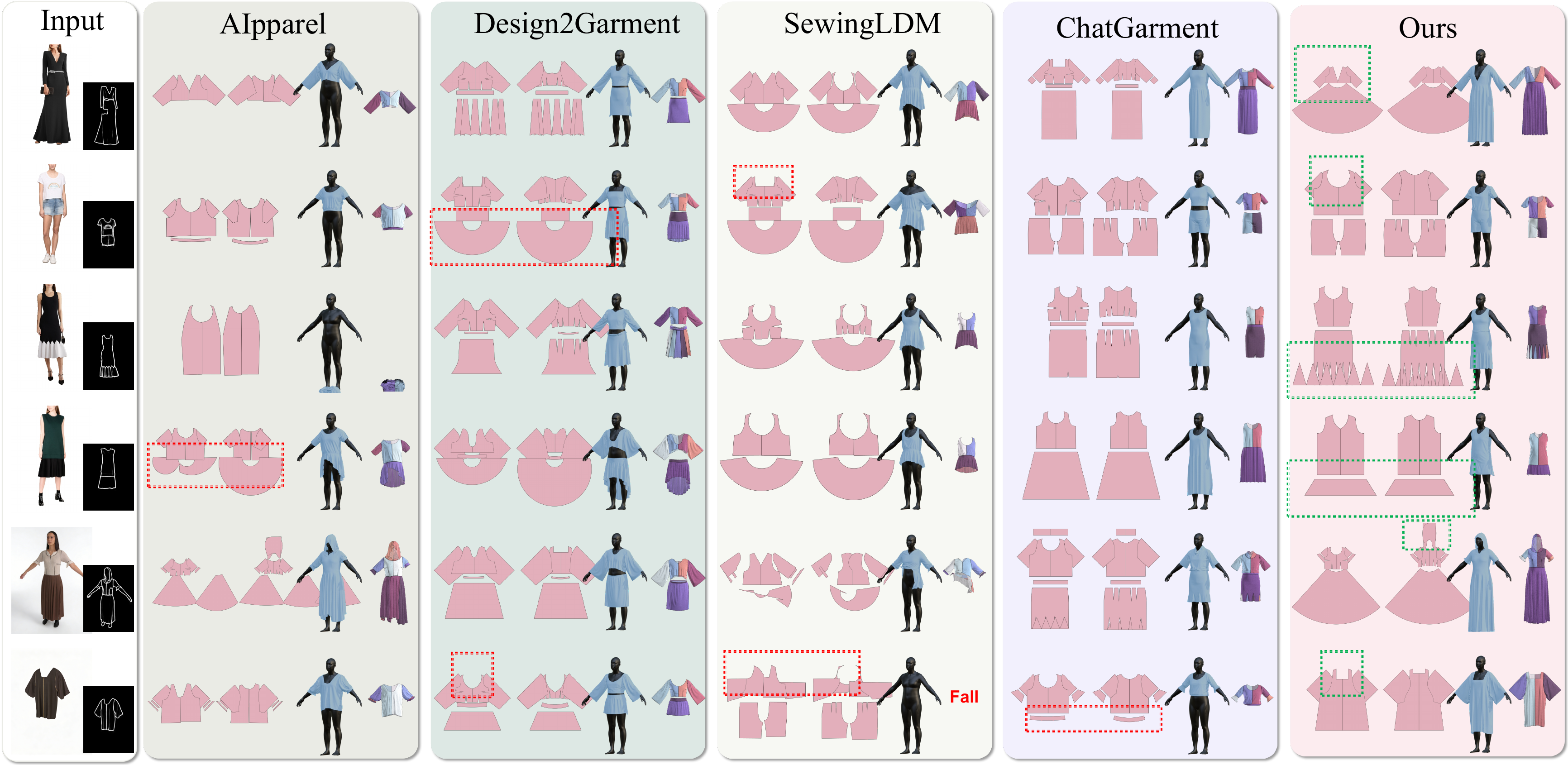}
  \caption{Qualitative comparison of sketch-conditioned Sewing pattern generation. From left to right are the input image and sketch, results of AIpparel, Design2Garment, SewingLDM, ChatGarment, and our method. GarmentWeaver better recovers garment categories and fine structural details while maintaining valid patterns for simulation. Green boxes highlight representative improvements, while SewingLDM may produce invalid patterns and Design2Garment may predict incorrect garment types.}
  \Description{Qualitative comparison of sketch-conditioned Sewing pattern generation results.}
  \label{fig:3}
\end{figure*}

% \section{Experimental Results}

\subsection{Experimental Setup}

\textbf{Dataset.}
Experiments are conducted on a large-scale garment dataset derived from GarmentCodeData \cite{7} and related multimodal garment design resources \cite{27,54,55}, containing approximately 120,000 samples spanning diverse garment categories and body-shape variations. Each sample is paired with a structured garment specification and two auxiliary modalities: a textual description generated by Qwen2-VL \cite{34} and a sketch extracted from garment images using PiDiNet \cite{36}. As a result, each sample is represented as a multimodal triplet of sketch, text, and structured garment target.

\noindent\textbf{Implementation Details.}
GarmentWeaver is built on LLaVA-7B \cite{32} and fine-tuned with LoRA \cite{33}. Training is conducted on a single vGPU-32GB GPU for 3 epochs, using a learning rate of $2\times10^{-5}$, a global batch size of 128, a maximum sequence length of 2048, and weight decay of 0. On the single-GPU setup, the effective batch size is maintained through gradient accumulation. During training, the model predicts executable garment programs in the proposed two-stage manner. To improve robustness under incomplete inputs, we independently drop the textual description and sketch with probability 30\%. The generated outputs can be converted into sewing patterns through GarmentCode \cite{6}. The full training process requires only 32 GB of GPU memory, and inference is efficient since the model directly predicts a compact structured output without iterative sampling.

\noindent\textbf{Evaluation Metrics.}
Following prior works \cite{23,24,25,26,3,8}, we evaluate predicted Sewing patterns using four metrics: Panel IoU, Trans L2, \#Panel, and \#Edge. Panel IoU measures the IoU between predicted and ground-truth 2D garment panels, and Trans L2 measures the L2 distance between predicted and ground-truth panel translations. \#Panel and \#Edge evaluate whether the predicted numbers of panels and panel edges match the ground truth. Higher is better for Panel IoU, \#Panel, and \#Edge, while lower is better for Trans L2.Together, these metrics provide a comprehensive evaluation of the generated sewing patterns in terms of geometric accuracy, spatial alignment, and structural fidelity.

\subsection{Qualitative Comparison}
\begin{figure*}[t]
  \centering
  \includegraphics[width=0.7\textwidth]{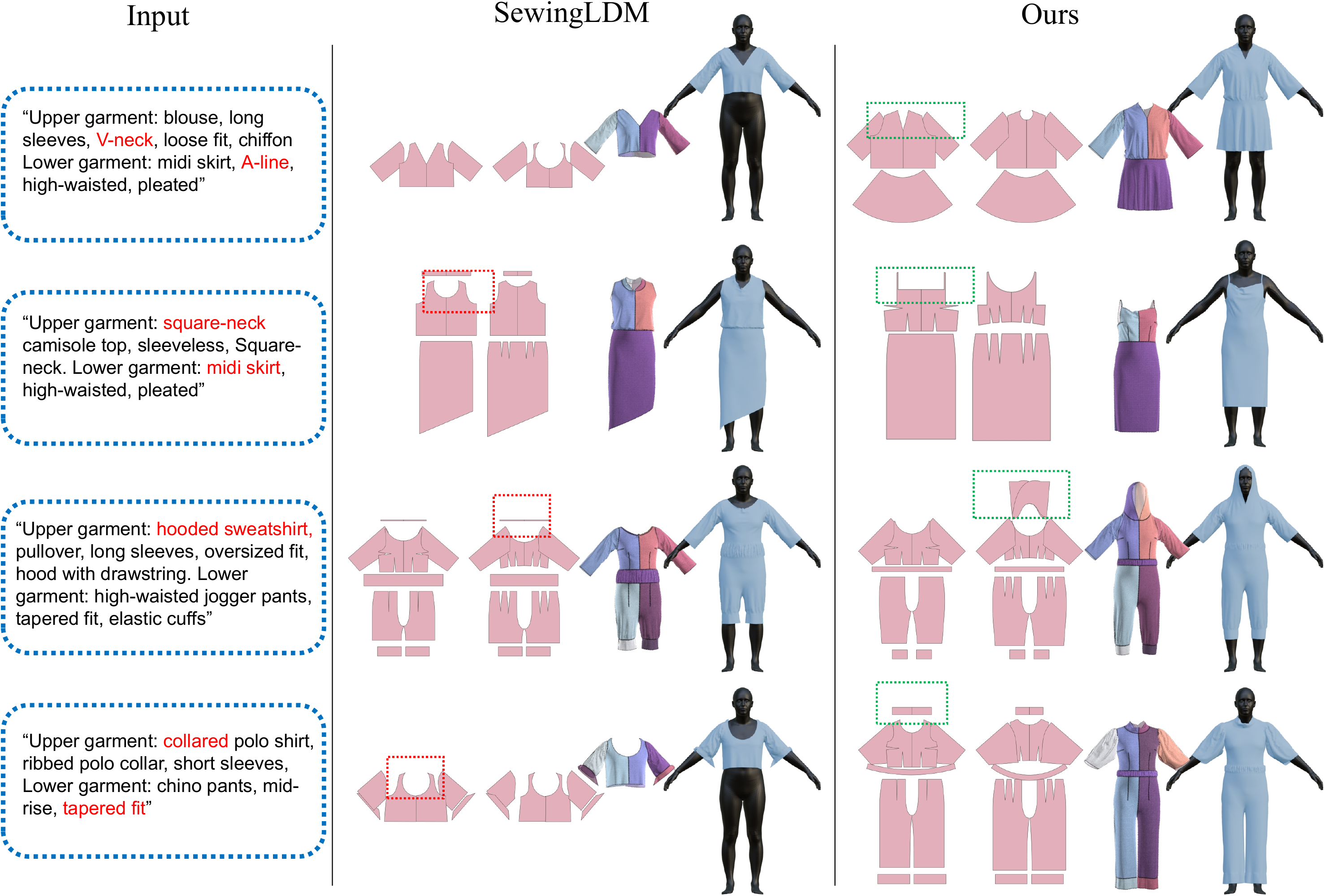}
  \caption{Qualitative comparison of text-only Sewing pattern generation. From left to right are the input descriptions, the results of SewingLDM, and the results of our method. GarmentWeaver better recovers key structural details, such as V-necks, square necklines, hoods, and collars, while producing more complete and semantically consistent garment parts. Green boxes highlight representative improvements.}
  \label{fig:4}
\end{figure*}
We compare GarmentWeaver with representative baselines, including AIpparel \cite{1}, Design2Garment \cite{3}, SewingLDM \cite{4}, and ChatGarment \cite{2}. As shown in Fig.~\ref{fig:3}, GarmentWeaver generates sewing patterns that are more consistent with the input sketches in both garment category and fine-grained structural details. Our method better preserves key design elements such as sleeve shape, bodice structure, and skirt proportion, while also producing patterns that support stable downstream simulation.

By comparison, Design2Garment sometimes predicts mismatched garment types, leading to semantic inconsistency with the input design. SewingLDM is less stable and may generate geometrically irregular or incomplete panels that are unsuitable for simulation. AIpparel and ChatGarment can recover coarse garment layouts, but often miss local structural details or produce oversimplified results. Overall, GarmentWeaver achieves a better balance between structural correctness, detail preservation, and simulation feasibility.

We also evaluate the text-only setting, where Sewing patterns are generated solely from natural-language descriptions. As shown in Fig.~\ref{fig:4}, compared with SewingLDM, GarmentWeaver more accurately recovers key text-specified components, especially distinctive structures such as V-necks, square necklines, hoods, and collars. It also produces pattern decompositions that better match the described garment structure, yielding more appropriate upper--lower combinations and more complete part layouts. In contrast, SewingLDM often captures only coarse garment appearance while missing important structural details or generating inconsistent component configurations.

Taken together, the qualitative results in both sketch-conditioned and text-only settings show that GarmentWeaver better preserves garment semantics, recovers finer structural components, and produces more executable sewing patterns than existing baselines.

\subsection{Quantitative Comparison}

\begin{table}[t]
  \centering
  \small
  \caption{Quantitative comparison with AIpparel, Design2Garment, SewingLDM, and ChatGarment on structured Sewing pattern generation. We report Panel IoU, Trans L2, \#Panel, and \#Edge. Higher is better for Panel IoU, \#Panel, and \#Edge, while lower is better for Trans L2. Our method achieves the best performance on Panel IoU, Trans L2, and \#Edge, demonstrating superior accuracy in panel reconstruction, spatial alignment, and structural detail recovery.}
  \label{tab:main_quant}
  \begin{tabular}{lcccc}
    \toprule
    Method & IoU$\uparrow$ & TransL2$\downarrow$ & \#Panel$\uparrow$ & \#Edge$\uparrow$ \\
    \midrule
    AIpparel & 0.834 & 1.783 & 91.3\% & 86.53\% \\
    Design2Garment & 0.852 & 1.169 & 94.5\% & 88.36\% \\
    SewingLDM & 0.793 & 1.200 & 97.8\% & 82.71\% \\
    ChatGarment & 0.864 & 1.147 & 98.4\% & 89.03\% \\
    Ours & 0.869 & 1.130 & 97.2\% & 89.52\% \\
    \bottomrule
  \end{tabular}
\end{table}

Table~\ref{tab:main_quant} compares GarmentWeaver with AIpparel \cite{1}, Design2Garment \cite{3}, SewingLDM \cite{4}, and ChatGarment \cite{2}. Our method achieves the best Panel IoU (0.869) and the lowest Trans L2 (1.130), demonstrating superior performance in both panel reconstruction and spatial alignment. In addition, GarmentWeaver obtains the highest \#Edge accuracy (89.52\%), indicating stronger capability in recovering fine-grained structural details of garment panels.

Although ChatGarment achieves the highest \#Panel accuracy (98.4\%), its geometric accuracy remains lower than ours, as reflected by inferior Panel IoU and Trans L2 values. Similarly, Design2Garment and SewingLDM show competitive performance on some structural metrics but remain less accurate in overall panel reconstruction. These results suggest that GarmentWeaver is more effective in generating structurally valid and geometrically accurate Sewing patterns. This observation is also consistent with the qualitative comparisons, where our generated patterns better preserve garment details and support stable simulation.

\begin{figure*}[h]
  \centering
  \includegraphics[width=0.9\textwidth]{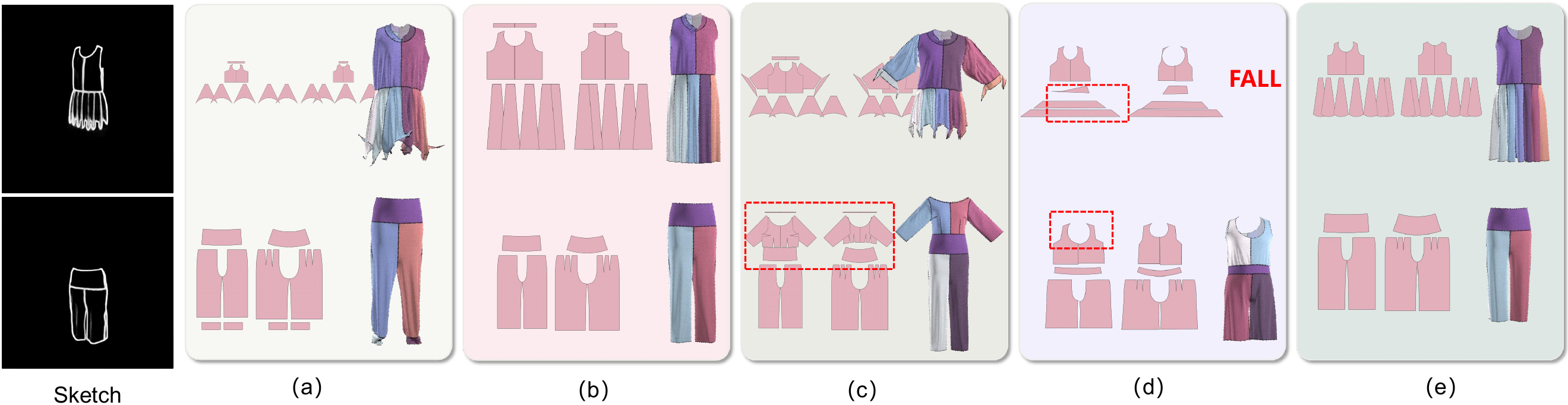}
  \caption{Qualitative ablation study of the proposed framework. From left to right: input sketch and the results of (a) w/o sketch guidance, (b) use LLaVA-13B backbone, (c) w/o feasibility-aware regularization, (d) direct full-sequence prediction, and (e) full model. Removing sketch guidance leads to correct garment categories but inaccurate shapes; removing feasibility-aware regularization introduces redundant structures; direct full-sequence prediction produces distorted and non-simulatable patterns. In contrast, the full model generates more regular sewing patterns and more reliable simulation results.}
  \label{fig:5}
\end{figure*}

\begin{table}[t]
  \centering
  \small
  \caption{Compactness analysis of the proposed schema-aware dynamic encoding. We compare the full-schema target length with the effective target length after schema-aware pruning under different garment structures.}
  \label{tab:compactness}
  \begin{tabular}{lccc}
    \toprule
    Garment Structure & Full Length & Effective Length & Reduction Ratio \\
    \midrule
    Overall & 118.24 & 59.75 & 49.53\% \\
    Upper only & 117.72 & 72.08 & 38.76\% \\
    Bottom only & 117.88 & 15.32 & 87.01\% \\
    Upper + Pants & 118.75 & 83.05 & 29.37\% \\
    Upper + Skirt & 118.76 & 82.87 & 30.01\% \\
    \bottomrule
  \end{tabular}
\end{table}

We further analyze the compactness of the proposed schema-aware dynamic encoding in Table~\ref{tab:compactness}. By removing inactive branches and preserving only structure-valid slots, the average target length is reduced from 118.24 to 59.75, corresponding to a 49.53\% reduction in target size. This compactness is beneficial in several aspects. First, pruning inactive branches directly reduces the number of parameters that need to be predicted, making the target representation more efficient than full-schema supervision. Second, full-schema prediction forces the model to allocate probability mass to many structure-irrelevant slots, which increases output ambiguity and makes autoregressive generation more prone to redundant or hallucinated components. In contrast, by restricting prediction to structure-valid branches only, the proposed encoding narrows the search space and allows the model to focus on the parameters that are actually relevant to the current garment. This more compact formulation also leads to a cleaner supervision signal, since the model is no longer distracted by inactive fields that carry no valid semantic content. As a result, the proposed encoding improves not only representation efficiency but also the effectiveness of learning valid structural dependencies for garment generation.

The reduction is also structure-dependent: bottom-only garments show the largest compression ratio because most upper-body branches in the full schema remain inactive and can be removed entirely. By contrast, garments containing both upper and lower parts retain a larger portion of the schema and therefore exhibit smaller reductions. These results confirm that the proposed representation not only improves structural validity but also reduces output redundancy.

\subsection{Ablation Study}

To analyze the contribution of each component, we conduct ablation studies on four variants: (1) w/o sketch guidance, (2) LLaVA-13B backbone, (3) w/o feasibility-aware regularization, and (4) direct full-sequence prediction instead of the proposed two-stage structured generation. The qualitative results are shown in Fig.~\ref{fig:5}.

Removing sketch guidance mainly degrades the shape accuracy of the generated patterns. The model can still predict the correct garment category, but fails to recover the target silhouette and structural proportions, indicating that sketch input is crucial for recovering contour-sensitive garment geometry. Replacing LLaVA-7B with LLaVA-13B does not lead to clear qualitative improvement, suggesting that the gain of GarmentWeaver does not primarily come from increasing backbone size, but from the proposed schema-aware target construction and two-stage structured generation.

When feasibility-aware regularization is removed, the model tends to generate redundant garment structures that do not exist in the target design. This verifies that the regularization term is important for suppressing invalid branch activations and improving structural coherence. A more severe degradation is observed when the proposed two-stage structured generation is replaced with direct full-sequence prediction. In this setting, the generated panels become distorted and irregular, and some results cannot be successfully simulated. In contrast, the proposed two-stage formulation produces cleaner, more standardized sewing patterns that are structurally valid and better suited for downstream simulation.

Overall, the ablation results confirm that GarmentWeaver benefits from both schema-aware target construction and feasibility-aware regularization. Sketch guidance improves shape fidelity, feasibility-aware regularization suppresses redundant components, and two-stage structured generation is essential for producing regular and simulation-compatible Sewing patterns.

\begin{table}
  \centering
  \caption{Comparison of Simulation Success Rate. For each method, the generated Sewing patterns are converted and simulated using the GarmentCode simulation pipeline. We report the percentage of samples that can be successfully simulated in 3D without failure.}
  \label{tab:sim_success_rate}
  \begin{tabular}{lc}
    \toprule
    Method & Simulation Success Rate (\%) $\uparrow$ \\
    \midrule
    AIpparel & 73.9 \\
    Design2Garment & 87.1 \\
    SewingLDM & 61.4 \\
    ChatGarment & 97.9 \\
    Ours & 98.3 \\
    \bottomrule
  \end{tabular}
\end{table}

We additionally evaluate Simulation Success Rate to measure the executability of generated sewing patterns. For each method, we randomly sample 1,000 results and perform 3D simulation using the GarmentCode simulator. As shown in Table~\ref{tab:sim_success_rate}, GarmentWeaver achieves the highest success rate (98.3\%), surpassing AIpparel (73.9\%), Design2Garment (87.1\%), SewingLDM (61.4\%), and ChatGarment (97.9\%). The relatively low success rates of SewingLDM and AIpparel are mainly related to their diffusion-based backbones. Methods built on diffusion models generate garment patterns in a continuous denoising space without explicit structural feasibility constraints, making them more prone to distorted panels, invalid part configurations, and unstable geometric details that lead to simulation failure. While ChatGarment also attains a high success rate, its flattened JSON formulation tends to favor component recall through memorization of frequent structural patterns, at the expense of parameter precision and geometric accuracy. By contrast, GarmentWeaver explicitly enforces schema-consistent structure and parameter validity, resulting in more reliable simulation-compatible outputs.

\section{Application}

\begin{figure}[b]
  \includegraphics[width=0.5\textwidth]{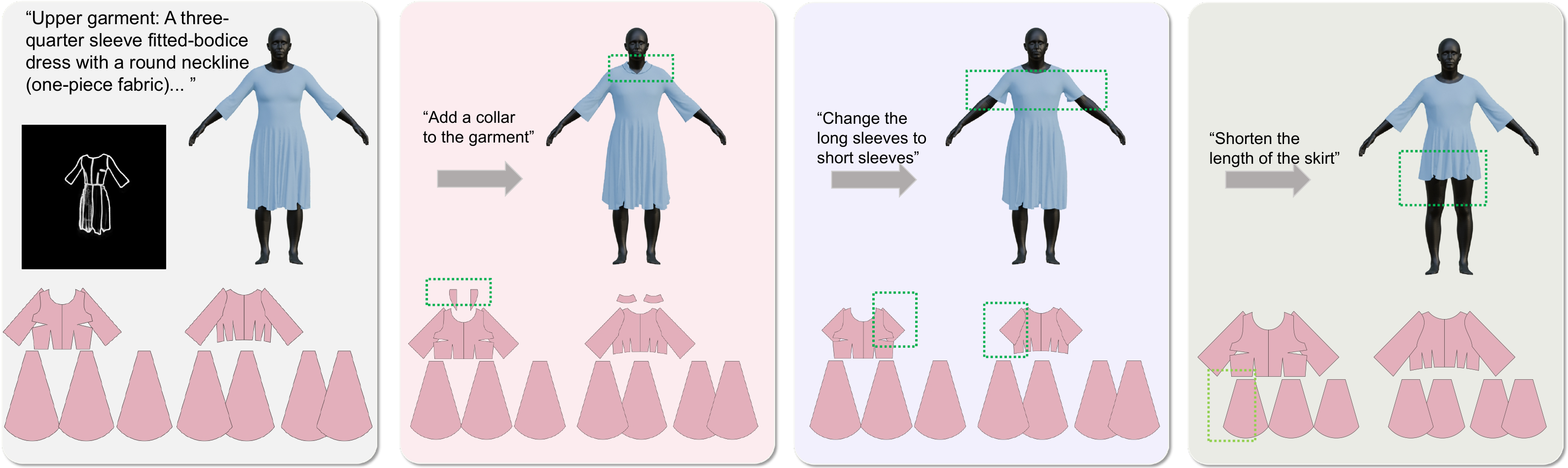}
  \caption{Instruction-based editing of sketch-generated Sewing patterns. Starting from an initial sewing pattern generated from the design input, GarmentWeaver progressively refines the pattern according to natural-language editing instructions, including adding a collar, changing long sleeves to short sleeves, and shortening the skirt length.}
  \Description{Instruction-based editing of sketch-generated Sewing patterns.}
  \label{fig:6}
\end{figure}

Beyond one-shot Sewing pattern generation, GarmentWeaver also supports instruction-based editing of sketch-generated patterns. Starting from an initial sewing pattern predicted from the design input, the model further refines the result according to a natural-language editing instruction. Thanks to the schema-aware structured representation, the editing process can be localized to the instruction-relevant components, so that only the affected structure or parameters are updated while unrelated garment parts remain unchanged. Fig. \ref{fig:6} shows several sequential editing examples, including adding a collar, changing long sleeves to short sleeves, and shortening the skirt length. These results demonstrate that GarmentWeaver enables controllable and localized refinement of generated sewing patterns, making it suitable for interactive garment design.

\section{Limitations and Future Work}

\begin{figure}[h]
  \includegraphics[width=0.4\textwidth]{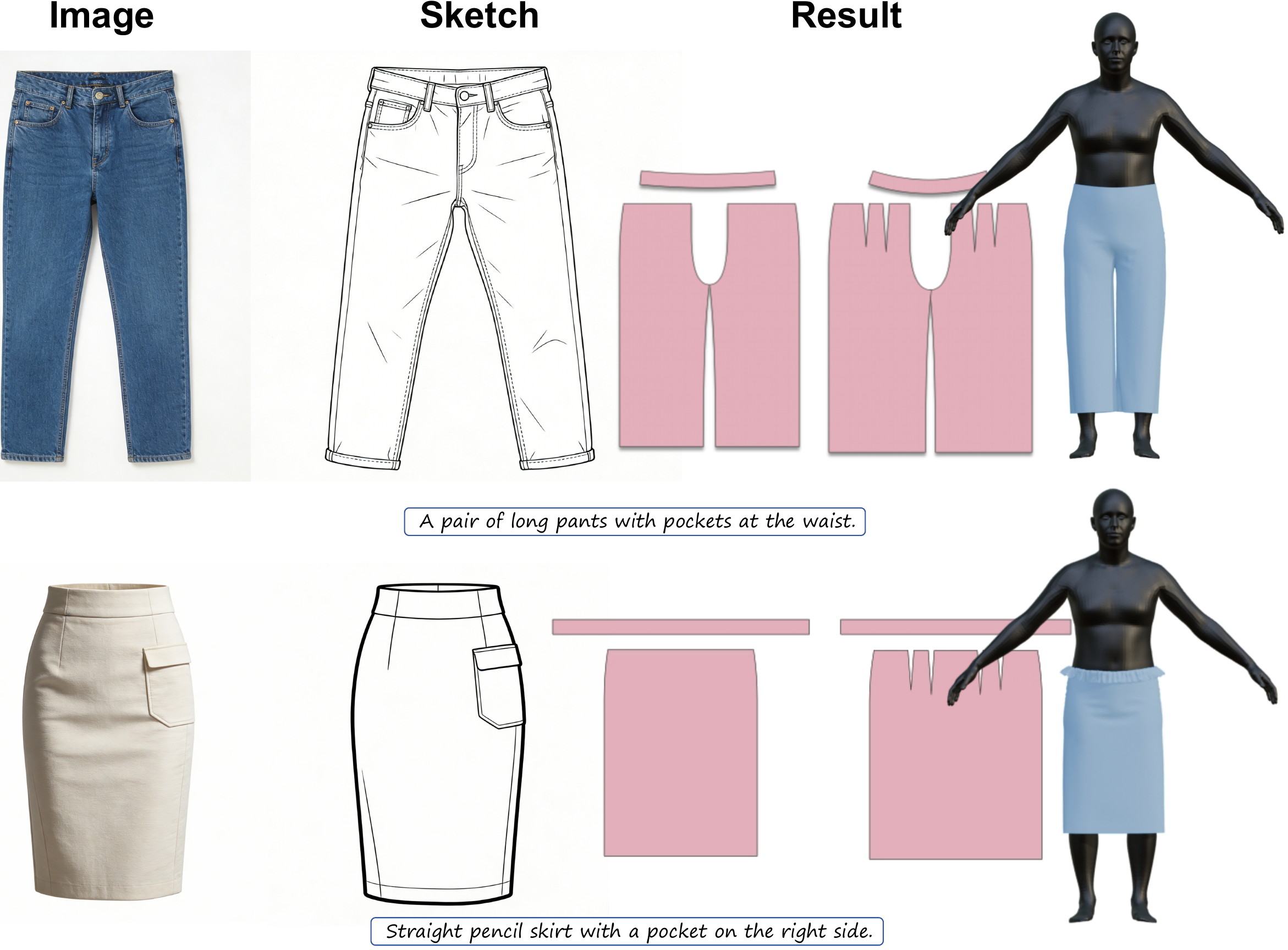}
  \caption{Limitations. For garments with localized accessory-like details, such as pockets, GarmentWeaver may fail to reconstruct the desired structures and instead only recover the overall pants or skirt shape.}
  \Description{Instruction-based editing of sketch-generated Sewing patterns.}
  \label{fig:7}
\end{figure}

Despite the promising results, GarmentWeaver still has several limitations. The main limitation comes from the underlying structured garment representation. Since our method relies on GarmentCode-style garment specifications, it is currently restricted to the garment components and parametric structures supported by that representation. Consequently, GarmentWeaver can generate the major silhouette and main structural parts of common garments, but it still struggles with finer localized details and accessory-like components that are not explicitly defined in the current schema, such as pockets, belt loops, and related attachments. As shown in Fig. \ref{fig:7}, even when pockets are clearly indicated in the reference image and sketch, the generated results mainly recover the overall garment shape while failing to reproduce the pocket structure. In future work, we plan to improve the bottom-level structure of GarmentCode by incorporating richer component definitions, especially for pockets and other localized attachments, so as to support more detailed multimodal garment generation.

\section{Conclusion}
We introduced GarmentWeaver, a schema-aware framework for multimodal sewing pattern generation. By modeling garment prediction in a more structured way, GarmentWeaver reduces the ambiguity of direct flat-sequence generation and improves the structural validity of the generated outputs. The proposed schema-aware target construction, structured generation process, and feasibility-aware regularization together enable more accurate, coherent, and simulation-compatible sewing patterns. Experimental results on both qualitative and quantitative evaluations demonstrate the effectiveness of our method over strong baselines.

\begin{acks}
To Robert, for the bagels and explaining CMYK and color spaces.
\end{acks}

\bibliographystyle{ACM-Reference-Format}
\bibliography{sample-base}

\end{document}